\documentclass[mnras]{llncs}
\usepackage[T1]{fontenc}
\usepackage{graphicx}

\usepackage{times}
\usepackage{soul}
\usepackage{url}
\usepackage[hidelinks]{hyperref}
\usepackage[utf8]{inputenc}
\usepackage[small]{caption}
\usepackage{graphicx}

\usepackage{booktabs}
\usepackage{algorithm}
\usepackage[switch]{lineno}
\usepackage{natbib}
\usepackage{algorithmic}
\usepackage{booktabs}
\usepackage[misc]{ifsym}
\usepackage{mathtools}

\usepackage{placeins}
\usepackage{mwe}

\begin{document}

\title{LGRPool: Hierarchical Graph Pooling Via Local-Global Regularisation}

\author{Farshad Noravesh\textsuperscript{1} \and
Reza Haffari\textsuperscript{2} \and
Layki Soon\textsuperscript{3} \and 
Arghya Pal\textsuperscript{4}
}

\institute{
Monash University, Malaysia \email{Farshad.Noravesh@monash.edu}
\and
Monash University, Australia \email{Gholamreza.Haffari@monash.edu}
\and
Monash University, Malaysia \email{soon.layki@monash.edu}
\and
Monash University, Malaysia \email{arghya.pal@monash.edu}
}
   
\maketitle              

\begin{abstract}
Hierarchical graph pooling(HGP) are designed to consider the fact that conventional graph neural networks(GNN) are inherently flat and are also not multiscale. However, most HGP methods suffer not only from lack of considering global topology of the graph and focusing on the feature learning aspect, but also they do not align local and global features since graphs should inherently be analyzed in a multiscale way.  LGRPool is proposed in the present paper as a HGP in the framework of expectation maximization in machine learning that aligns local and global aspects of message passing with each other using a regularizer to force the global topological information to be inline with the local message passing at different scales through the representations at different layers of HGP. Experimental results on some graph classification benchmarks show that it slightly outperforms some baselines.
\end{abstract}

\section{Introduction}
The modern approach to message passing in graph neural networks (GNNs) introduces a significant improvement by decoupling feature learning from message propagation. Traditionally, message passing tightly intertwined the two processes, where node features were updated directly based on aggregated information from neighbors. However, this approach often led to challenges like oversmoothing in deep networks and limited flexibility in processing complex structures. By decoupling, feature learning is handled independently using techniques like learnable transformations \cite{EliChien2020} and \cite{Wimalawarne2021} while message propagation focuses solely on distributing and aggregating information across the graph. This separation enhances model expressiveness, as feature learning can leverage advanced techniques tailored to the data, while propagation dynamics can be optimized for the graph's topology. Consequently, this approach leads to more robust and scalable GNNs, with improved performance in tasks like link prediction, node classification, and graph-level representation learning.
\par
Multiscale graph representation and global topological features are integral to understanding the intricate structure and dynamics of complex systems. Multiscale graph representation allows for the analysis of graph across varying levels of detail, capturing both local interactions and overarching structural patterns which is vital for graph classification tasks. This hierarchical approach facilitates the exploration of high level graph structure without losing essential finer details. Simultaneously, incorporating global topological features—enables a deeper understanding of the overall shape and behavior of the system.
\par
As graphs often contain complex and high-dimensional, directly analyzing the entire graph at once can be computationally expensive and prone to many problems such as overfitting, oversmoothing and oversquashing. Hierarchical pooling addresses this by progressively reducing the graph's size through cluster or node selection methods, thereby summarizing local structures into coarser representations. This allows models to capture both global and local patterns more effectively, facilitating tasks like graph classification, regression, or node prediction. Most GNNs employ a spatial operator based on graph laplacian which limits the radius of receptive field in a Graph. Defining a general convolution operator in the graph domain is challenging due to the lack of canonical coordinates \cite{LihengMa2024},\cite{MosheEliasof2022}. In contrast to conventional message passing methods for GNN in the literature, a flexible message passing paradigm for GNN may involve a layer called pooling. Pooling is not as straightforward as pooling in computer vision since graph could not be reduced to a grid as is the case for images. 
\par
There are different paradigms for graph pooling in the literature. However, the global topological features are either mixed with feature learning and attributes or they are not modeled in a multiscale way. The multiscale information could be easily modeled by hierarchical graph pooling. These two objectives namely capturing global topological graph information such as centrality on the one hand, and representing graph using multiscale modeling are separately analyzed in the literature. In this paper, global topological features are modeled by personalized page rank in the propagation step and are compared with the last hierarchical pooling layer to align them via a regularizer and to enforce their difference as small as possible. The proposed method is formulated as an expectation maximization problem. In the expectation step, the goal is to separate feature learning from message propagation that can capture multihop information. Once a good latent representation of nodes are obtained, the maximisation step adjusts these representation with multiscale nature of graph through a regularization term. Note that LGRPool should be seen as a framework and any submodule could be implemented differently. For example we used edgePool for hierarchical pooling since there is no constraint on the number of clusters in advance which makes the model more adaptive to graph dataset distribution. Since the major goal of LGRPool is connecting global topological information  of nodes with multiscale nature of a graph, any implementation such as using DiffPool \cite{RexYing2018} could be considered for further improvement of the performance.

The following are three major contributions of the present paper:
\begin{itemize}
\item {Designing and developing an expectation maximization framework that considers the feature vector as a latent variable and through a regulariser aligns local features to global topological features.  
 }
 \item{The present proposed framework has the property that considers feature learning, propagation and multiscale nature of graph classification datasets by decoupling different objectives.}
 \item{Experiments on four datasets on some graph classification benchmarks shows that our method slightly outperforms SOTA baselines on two datasets.}
\end{itemize}
\begin{figure*}[ht]
\centering
\includegraphics[scale=0.3]{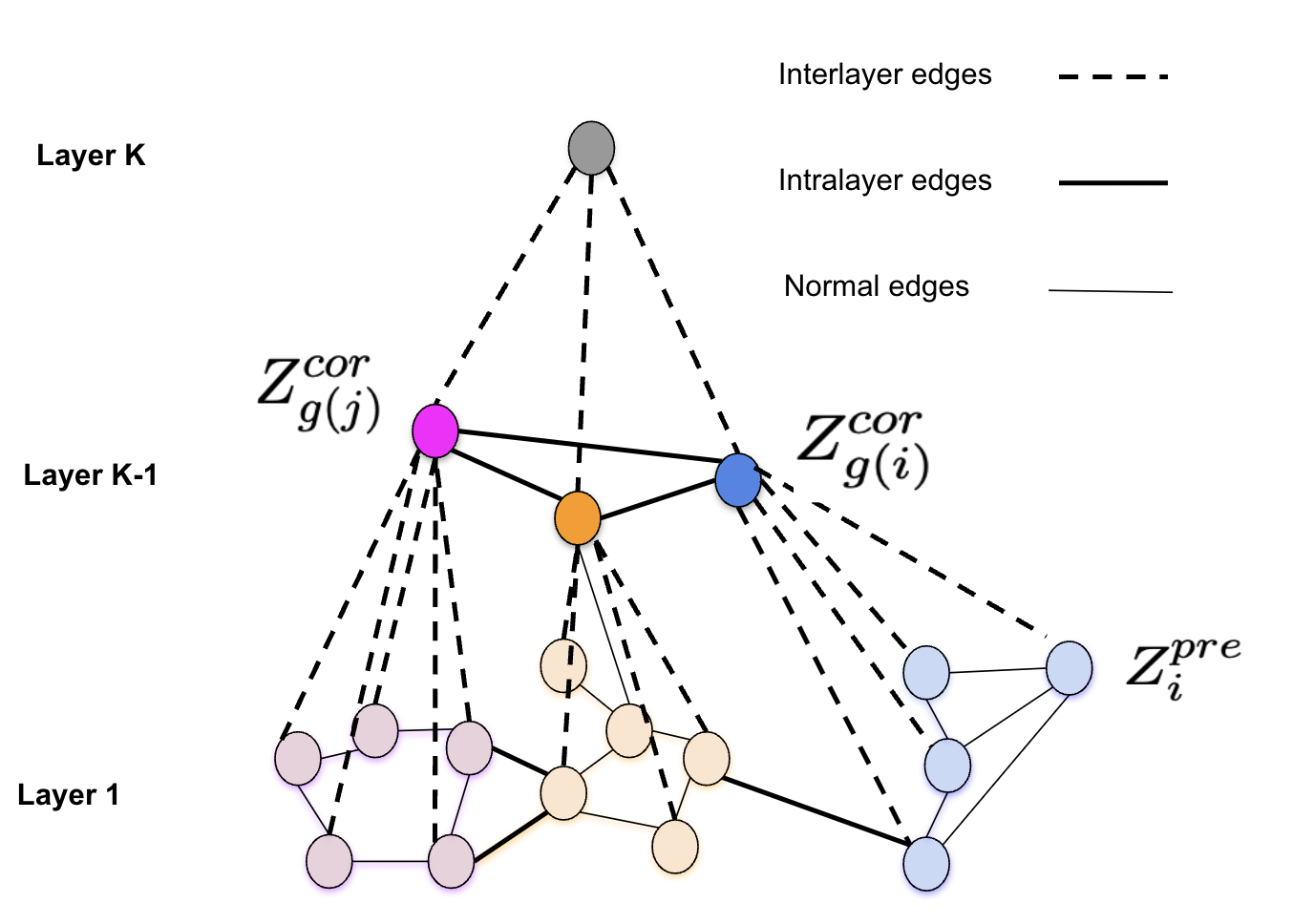}
\caption{local global message passing}
\label{fig-anealingProcess.png}
\end{figure*}
\begin{figure*}[ht]
\centering
\includegraphics[scale=0.30]{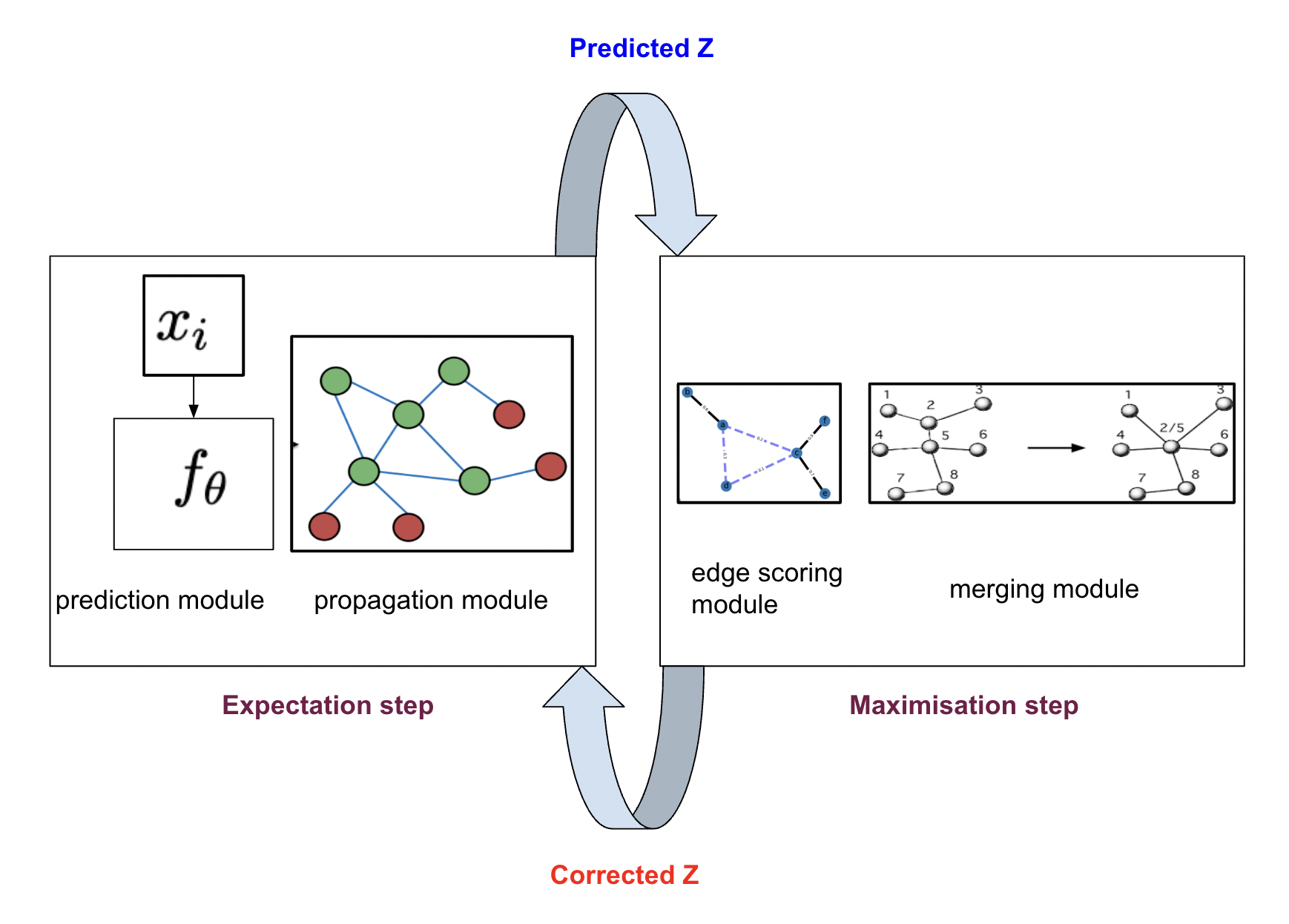}
\caption{proposed architecture}
\label{fig-proposed}
\end{figure*}
\section{Related Works}

\subsection{Graph Pooling}\label{graphpoolingmethods}
Pooling for graphs could be classified into the following four categories:
\begin{enumerate}
\item{Global Pooling: One of the most effective types of global pooling methods is a Multiset Encoding method called GMPool in \cite{JinheonBaek2021} that captures the interaction between nodes according to their structural dependencies.  Multiset Encoding allows for possibly repeating elements, since a graph may have redundant node representations and leverages an idea similar to set transformers \cite{JuhoLee2018} to model node interactions and to compress n nodes into k typical nodes. Unfortunately, most global pooling methods use a permutation invariant function such as summation or maximum that totally ignores the information inherent in global topology of a graph and the local substructures and the high order graph between substructures. Thus, it is expected that the performance on graph classification benchmarks to be very poor as experiments confirm this expectation.
 }
\item{Cluster identification: These methods compute the dense cluster assignment matrix with an adjacency matrix. This prevents them from exploiting sparsity in the graph topology. This is usually done by projecting node features on a learned weight to obtain an assignment matrix. Nodes that have close embeddings are projected on the same cluster. After having obtained the assignment matrix, super nodes at the coarsened level can be computed by aggregating all nodes that belong to the same cluster \cite{RexYing2018}.
There are soft and hard approaches to cluster assignment. SSHPool proposed in \cite{XuZhuo2024} is an example of hard assignment that each node can belong to only one cluster and is obtained by doing hard assignment over a soft prediction. ABDPool \cite{YueLiu2022} is another example of hard assignment which is done by an attention mechanism. Although \cite{Bianchi2019} leverage relaxed formulation of mincut, this approach is still in the category of cluster assignment identification. In contrast to mincut objective, \cite{Tsitsulin2024} leverages modularity objective which has shown that it has better performance and easier training process specially in larger graphs in comparison to \cite{Bianchi2019} which only uses mincut objective. All methods in this category need regularizers to be effective as is shown in Table~\ref{tab:methods}. 
}
\item{Top-k nodes approach:  Like \cite{HongyangGao2019-gpool},\cite{HongyangGao2019-graphUnet}, the objective is to score nodes according to their importance in the graph and then to keep only nodes with the top-k scores. Node drop methods unnecessarily drop some nodes at every pooling step, leading to information loss on those discarded nodes. One drawback of this approach is that the reduced graph in each pooling layer might be end up with a discontinuous graph or it may ignore the local substructures. To address this issue, \cite{Stanovic2025} introduces MISPool which uses maximal independent sets(MIS) to ensure that the pooled graph at each layer is connected. They name these selected nodes as survival nodes that are obtained by MIS algorithm. Other approaches preserve connectedness of the graph such as \cite{XuMingxing2022} that introduced Liftpool which has similar performance to SAGPool in \cite{JunhyunLee2019}, since both of them use the same node scoring method. Liftpool ignores the topology of the graph and uses a feature map which is obtained by conventional local message passing methods. The methods discussed so far do not leverage the global topology of the graph. Thus ENADPool is proposed in \cite{ZhehanZhao2024} that simultaneously identifies the importance of different nodes within each separated cluster and edges between corresponding clusters. The global topology is encoded by masking the generalized graph diffusion(GGD). It employs a hard clustering strategy to assign each node into a unique cluster.
}
\item{Edge based pooling: An edge contraction pooling layer has recently been proposed by \cite{Diehl2019}. They compute edge scores in order to successively contract pairs of nodes, which means that they successively merge pairs of nodes that are linked by edges of the highest scores. With the same analogy, \cite{Snelleman2024} takes any two nodes by a simple linear layer followed by a nonlinearity and merges them if the value is bigger than a threshold. With the same spirit, \cite{Galland2021} introduces a more general approach for edge based pooling and adds a regularization term to include the normalized cut between clusters. One problem of edgepool is that the quota in each layer is fixed. To circumvent this issue, \cite{WuJunran2022} introduces SEP which uses structural entropy to guide merging of a set of nodes instead of merging just two nodes as is the case in edgepool. Although the order of complexity of SEP is linear in the number of edges, it is not clear how structural entropy could be an appropriate measure to respect the global topology of the graph and how it avoids producing local structure damage.
}

\end{enumerate}

Node drop methods unnecessarily drop arbitrary nodes, and node clustering methods like DiffPool have limited scalability to large graphs. \cite{RexYing2018} was one of the first GNN based approaches to graph pooling that suffers from single nonconvex graph classification objective. Thus, link prediction and entropy of clusters were added to it to make it easier to train. However, it still suffers from huge computational problems and lacks theoretical foundations. To address this issue, \cite{Bianchi2020} relaxed the classical k-way cut problem which is NP-Hard and added auxiliary orthogonality constraint as is shown in Table~\ref{tab:methods}. \cite{Tsitsulin2024} showed experimentally that the objective of MunCutPool is not easy to train in huge graphs like social media datasets. Thus, they introduced DMoN which maximises the popular modularity objective in community detection literature.

\begin{table*}[h!]
    \centering
        \caption{state-of-the-art models for hierarchical pooling in the category of cluster identification}
        \label{tab:methods}
        \resizebox{0.9\textwidth}{!}
        {
        \begin{tabular}{p{3cm}|p{2cm}|p{3cm}|p{3cm}|p{4cm}}
            {\textbf{Author}}  & \textbf{model name} & \textbf{main objective} & \textbf{auxillary objectives} & \textbf{cluster assignment} \\
            \hline
            \text{\cite{RexYing2018}} & \textbf{DIFFPool} & \textbf{graph classification} & $ L_{LP}+\frac{1}{n}\sum_{i=1}^{n}H(S_{i})$ & $S=\textbf{softmax}GNN_{l}(A^{l},X^{l})$ \\
            \text{\cite{Bianchi2020}}  & \textbf{MinCutPool} & $-\frac{Tr(S^{T}AS)}{S^{T}DS}$ & $||\frac{S^{T}S}{||S^{T}S ||_{F}}-\frac{I_{K}}{\sqrt{K}}||_{F}$ & $S=MLP(X;\theta)$ \\
            \text{\cite{Tsitsulin2024}} & \textbf{DMoN} & $-\frac{1}{2m}Tr(C^{T}BC)$ & $\frac{\sqrt{k}}{n}||\sum_{i}C_{i}^{T}||_{F}-1 $ & $C=\textbf{softmax} GCN(A,X)$ \\
            \text{\cite{Bhowmick2024}}  & \textbf{DGCluster} & $-\frac{1}{2m}Tr(BXX^{T})$ & $\frac{1}{|S|^{2}}||H-X_{S}X_{S}^{T} ||_{F}^{2}$ &  \textbf{k-means of transformed X}  \\
        \end{tabular}
        }
\end{table*}

Methods like DiffPool and MinCutPool still have time and space complexity problems mainly due to cluster assignment matrix computation \cite{Haddadian2024}. Node dropping methods use scoring functions to locate just a subset of nodes that have high scores. While TopK \cite{HongyangGao2019} completely ignores the graph topology during pooling, SAGPool and gPool \cite{HongyangGao2019-gpool} modify the TopK formulation by incorporating the graph structure. The novelty of SAGPool is introduction of self-attention score that uses an activation function like tanh and a top-rank function that returns the indices of the top values. Both TopK and SAGPool avoid computing the cluster assignment matrix which reduces computational complexity. Unlike previous methods for node scoring, \cite{Haddadian2024} introduces MagPool that leverages personalised pagerank for feature propagation similar to \cite{Bojchevski2020}. DiffPool requires space complexity of $O(k|V|^{2})$ while gPool has requires only $O(|V|+|E|)$ which is a big improvement in terms of space complexity. Methods like DiffPool also need several auxiliary objectives like link prediction and cluster assignment entropy regularization to train well. 
\par
Since attention score of each edge is also important, \cite{Haddadian2024} introduces a framework for hierarchical pooing in which each pooling layer has a sequential architecture. The first stage is attention layer which scores each edge locally and neglecting graph topology. The second stage, focuses on multihop attention and the objective is topological. Thus, here it calculates personalized pagerank iteratively and then use it for message propagation. Finally, the last layer is pooling that scores the top K nodes as follows:
\begin{equation}\label{edge-scoring-magpool}
\begin{split}
S^{l}&=\sigma(A_{K}H^{l}W^{(l)}_{a})  \\
idx&=TopK(S^{(l)},[rN])
\end{split}
\end{equation}
where $W_{a}$ is a trainable vector to aggregate approximated information into node scores and $\sigma$ is a tanh nonlinearity.
\subsection{Structural Similarity}
Structural information of the graph, typically in the form of Laplacian eigenvectors or random walk transition probabilities are necessary since the conventional message passing methods which involve aggregating information in the 1-hop neighborhood prevent the model from learning coarse topological structures. It is important to emphasize that the phrase "structural similarity"(SS) is purely a local topological measure like when two nodes are part of a clique, they have similar topological roles and it could be scored recursively based on the similarity of their neighbours as is defined in \cite{YuZhizhi2024}. Another important structural properties are such as coreness that could be used to construct graph kernels \cite{Kalofolias2021}. Structural properties of two distant nodes could be the same if for example their neighborhoods has a special clique or triangle. 
\par
The local information and global information in \cite{YuZhizhi2024} are combined under the framework of adaptive graph convolutional networks. Although the local information matrix and the global information matrix are defined separately, they are added together to define the representation of each node and the alignment is lost during this process. \cite{Eijkelboom2023} uses a tensor product of features and structure and is experimentally shown to be more effective than the concatenation of the two. \cite{DexiongChen2020} leverages kernel and Nystrom approximation for node embedding based on random walks but the k-means algorithm and preprocessing makes it computationally expensive. \cite{QingqingLong2021} follows the same idea of \cite{DexiongChen2020} but adds extra feature which is derived from anonymous random walk. With the same spirit, \cite{AosongFeng2022} leverages kernel that mimics the same analogy of convolutional networks and each filter has a trainable adjacency matrix and unlike the set of random walks, is not invariant to any permutation and the learned filters are based on a particular permutation. The main research gap among articles like \cite{QingqingLong2021} , \cite{DexiongChen2020},\cite{AosongFeng2022} is the lack of an inductive bias due to global topology which could be modelled by any special case in generalized graph diffusion(GGD). \cite{IsaacReid2023} resolves this gap by considering GGD as a gram matrix of a graph kernel function. The structure information in  \cite{Eijkelboom2023}  is simply the concatenation of random walks for different lengths. This approach combines different scales of structural information. In contrast, message passing in hierarchical graph pooling methods are done at different scales and structural information are encoded at different scales which justifies why most hierarchical approaches outperforms the conventional methods. This multiscale nature of structures is one of the motivations of the present work.
\par
In the hierarchical graph pooling(HGP) framework like SSHPool, this local-global information is explicitly achieved in multiple pooling layers. Although SSHPool utilizes a graph attention layer to align the local information of samples subgraph with the global features, local-global alignment is still a serious challenge in such modelings. SEP uses structural information entropy for to consider local structures but the alignment of local information with global embedding is not obvious. While SEP does not distinguish between different structures, SPGP first captures and enumerates cliques or BCC and learns the score of each node to check if it belongs to a type of structure. The main drawback of such a local global alignment is that the structures are limited to two types namely BCC and cliques and an intensive preprocessing is required for such enumeration over all nodes of the graph. The present paper aligns the local and global information in a single representation by defining a local-global regulariser and is not limited to any type of structure such as BCC. 
\subsection{Generalized Graph Diffusion}
There are many methods that capture local or global topology of a graph such as positional encoding in \cite{Gabrielsson2022} that uses powers of adjacency matrix and is a special case of GGD. This power series can also be used to calculate the general random walk kernel(GRWK) in \cite{KrzysztofChoromanski2024}. Geometric random walk kernels and exponential kernels and popular kernels such as marginalized graph kernel will appear as some special cases of this GRWK. Two main families of node feature augmentation schemes exist for enhancing GNNs: random features and spectral positional encoding. Random Feature Propagation (RFP)\cite{MosheEliasof2023}, is inspired by the power iteration which has implicit relationship with GGD and the propagation matrix could either be learned from data or it can be predetermined using some powers of the adjacency matrix. 
Note that LGRPool does not use any feature augmentation schemes and the features are just the natural physical attributes. Topological features such as positional encoding could be concatenated to these original features in the future works.
Another methods capture global topology of a graph such as effective resistance(ER) in \cite{XuShen2024}. However, the calculation of ER requires computing pseudo inverse of laplacian. A special case of GGD is when the number of walks between two nodes of at most k is the only measure of connectivity as in \cite{Barbero2024}. \cite{IsaacReid2023} constructs a random feature map to provide an unbiased estimation of GGD using modulation function which upweights or downweights the contribution from different random walks depending on their lengths. Personalized PageRank(PPR) and heat kernel are just some special cases of GGD \cite{JohannesGasteiger2019} and are closely related to spectral based models originated from spectral graph theory.
\cite{Gasteiger2019} uses an adaptation of personalized pagerank(PPR) by the following recurrent equation:
\begin{equation}\label{eq:recursive_ppr}
\pi_{ppr}(i(x)) = (1-\alpha)\hat{\Tilde{A}}\pi_{ppr}(i_{x})+\alpha i_{x}
\end{equation}
where $i_{x}$ in \eqref{eq:recursive_ppr} is the teleport vector that allows us to preserve node's local neighborhood even in the limit distribution. The explicit solution of \eqref{eq:recursive_ppr} is as follows:
\begin{equation}
\Pi_{ppr}=\alpha (I_{n}-(1-\alpha)\hat{\Tilde{A}})^{-1}
\end{equation}
\cite{Roth2022} analyzes and encodes the effect of initial distribution on the performance of PPR.
\subsection{Decoupling Structure from Featurability}
Current deep GNN models entangle representation transformation and propagation and this hinders learning the graph node representations from larger receptive fields. These traditional deep GNNs have multiple layers which capture multiple hops and each layer has aggregation of node's neighbours. Nevertheless, one layer of these neighborhood aggregation methods only consider immediate neighbours, and the performance deteriorates when going deeper to enable larger receptive fields. Thus, many methods such as \cite{MengLiu2020} have been developed to address this issue by decoupling transformation from propagation. \cite{Nikolentzos2020} learns the hidden structures inside a graph in a differentiable way using different features related to different lengths of random walk but has the drawback that their kernel couples attribute information(features) from structural properties and makes training difficult since two different objectives are entangled with each other and can not be learned efficiently. Thus, an important research gap is to decouple aspects like local structure, attributes, and global topological features like positional encoding. 
\par Using PPR for GNN is discussed by many researchers such as \cite{Roth2022}. \cite{Roth2022} utilise nonlinear mapping and fixed point of the equilibrium condition as a way to leverage the stationary distribution and encoding topological structure of the graph. \cite{Gasteiger2019} introduced personalized propagation of neural predictions (PPNP) which decouples features in node prediction from message propagation using personalized PageRank. Thus,
the predicted node labels are as follows:
\begin{equation}\label{eq-PPNP}
\begin{split}
Z_{PPNP} &= \textbf{sotfmax} (\alpha(I_{n}-(1-\alpha)\hat{\Tilde{A}})^{-1}H) \\
H_{i,:} &= f_{\theta}(X_{i,:})
\end{split}
\end{equation}
It is obvious from \eqref{eq-PPNP} that the depth of the neural net is now independent of the message passing. Moreover, personalized PageRank can use even infinitely many layers which is impossible for classical message passing due to oversmoothing and oversquashing phenomena. Since directly calculating the inverse matrix in \eqref{eq-PPNP} is hard, some authors like \cite{Bojchevski2020} introduced approximations of the personalized pagerank.
Likewise, \cite{EliChien2020} and \cite{Wimalawarne2021} decoupled topology and node features using a generalized pagerank.
\par
\begin{table*}[h!]
\centering
\small 
\scalebox{0.9}{
\begin{tabular}{p{2cm}p{2cm}p{2cm}p{2cm}p{2cm}}
\hline
$\gamma$ & \textbf{MUTAG} & \textbf{Proteins} & \textbf{DD} & \textbf{NCI1} \\ \hline
0.10   & 69.91       & 70.62        & 74.17      & 68.83      \\ 
0.15   & 78.21      & 71.97 & 77.31      & 72.51      \\ 
0.20   & \textbf{82.18}  & 73.17   & \textbf{77.73}     & \textbf{75.45}     \\ 
0.25   & 79.73 & \textbf{73.19}  & 77.57 & 74.74      \\ 
0.30 & 75.38      & 72.71 &  76.65 &  62.74 \\ \hline
\end{tabular}
}
\caption{The effect of varying $\gamma$ on Graph classification accuracies on four benchmarks (percentage). }
\label{tab:gamma}
\end{table*}
Another approach to decouple structure from attributes can be done by concatenating the structural attributes such as k-step return time probabilities with the physical attributes and then embedding it in Hilbert space using implicit mappings and tensor product of kernels.
as is done in \cite{ZhenZhang2018}. The main drawback of such an approach is the global pooling that is done by summing features of all nodes to create mean embeddings which entangles different types of information and therefore reduces expressiveness of learning.
Thus, the need to decouple structural attributes from the original attributes is one of the motivations of LGRPool. It should be noted that LGRPool not only decouples structural attributes from original attributes, but also aligns the local and global node embeddings and achieves it in a hierarchical way.
\section{Problem Formulation}
\begin{algorithm}
  \begin{algorithmic}
    \STATE Input : (graph) from dataset \\
    \STATE 1: Loop until prediction-correction error is less than a threshold: \\
    \STATE 2: \textbf{Expectation Step}: \\
    \STATE 3: Train the propagation module with graph classification loss defined in Equation~\eqref{eq-expectationstepLoss} \\
    \STATE 4: Freeze the weights($\theta$) of propagation model \\
    \STATE 5: \textbf{Maximisation Step}: \\
    \STATE 6: Train with frozen weights of propagation module to obtain the edge score in Equation~\eqref{eq:edgeScoring} \\
    \STATE 7: cluster based on edge score thresholding and merge nodes in Equation~\eqref{eq:norm} \\
    \STATE 8: backpropagate the total loss defined in Equation~\eqref{eq:total} \\
    \STATE Output: $\theta$ and $W_{pool}$ and $a$  \\
    \caption{LGRPool algorithm for hierarchical graph pooling}
    \label{alg:t}
  \end{algorithmic}
\end{algorithm}
\begin{table*}[h!]
\centering
\scalebox{0.9}{
\begin{tabular}
{p{2cm}p{2cm}p{2cm}p{2cm}p{2cm}} \hline  
  dataset &  \textbf{MUTAG}  &   \textbf{Proteins} &  \textbf{DD} & \textbf{NCI1}  \\  
 graphs   &     188 &   1,113 &   1,178 &   4,110\\  
 classes &  2 &   2 &   2 &   2\\  
 average nodes   &  17.9 &   39.1 &   284.3 &    29.8 \\  \hline
\end{tabular}
}
\caption{bioinformatics dataset statistics}
\label{tab:dataset_statistics}
\end{table*}
The proposed architecture is shown in Figure~\ref{fig-proposed}. Hierarchical pooling on graphs could be seen as an expectation maximisation (EM) step in which the latent variables are node feature vectors. Algorithm~\ref{alg:t} shows the proposed EM method.
The expectation step consists of two modules namely prediction and the propagation module which provides an estimate of feature vector which is the latent variable in our framework. In the maximization step, graph classification objective is achieved through the trainable matrices in the edge scoring.
\begin{table*} [h!]
\centering
\small 
\scalebox{1.0}{ 
\begin{tabular}{p{2cm}|p{2cm}|p{2cm}|p{2cm}|p{2cm}}
\hline
\textbf{Model} & \textbf{MUTAG} & \textbf{Proteins} & \textbf{DD} & \textbf{NCI1} \\ \hline
TopKPool   & 67.61±3.36       & 70.48±1.01        & 73.63±0.55      & 67.02±2.25      \\ 
ASAP       & 77.83±1.49       & \underline{73.92±0.63} & 76.58±1.04      & 71.48±0.42      \\ 
SAGPool    & 73.67±4.28       & 71.56±1.49        & 74.72±0.82      & 67.45±1.11      \\ 
DiffPool   & \underline{79.22±1.02} & 73.03±1.00        & \underline{77.56±0.41} & 62.32±1.90      \\ 
MinCutPool & 79.17±1.64       & \textbf{74.72±0.48} & \textbf{78.22±0.54} & \underline{74.25±0.86} \\ 
LGRPool(ours)    & \textbf{81.56±1.53} & 73.51±0.63        & 77.51±0.67      & \textbf{75.45±0.52} \\ \hline
\end{tabular}
}
\caption{Graph classification accuracies on four benchmarks (percentage). The shown accuracies are mean and standard 
deviation over 10 different runs. We use \textbf{bold} to highlight wins and \underline{underline} to highlight the second best.}
\label{tab:graph_classification_results}
\end{table*}
\begin{table}[h!]
\centering
\begin{tabular}{|c|c|}
\hline
\textbf{hyperparameters} & \textbf{values} \\
\hline
\textbf{batch}  & 32 \\\hline
\textbf{num pooling layers} & 14 \\\hline
$k$ & 10 \\\hline
$\alpha$ & 0.3 \\\hline
\textbf{epochs} & 100 \\\hline
\textbf{hidden} & 200 \\\hline
\textbf{dynamic learning rate} & 1e-3 \\\hline
\textbf{optimizer} & Adam  \\\hline
\end{tabular}
\caption{hyperparameters}
\label{tab:hyperparameters}
\end{table}
 \par
The maximization step consists of edge scoring module and the merging module which could be implemented in different ways. LGRPool only uses edgepool \cite{Diehl2019} to merge the nodes after scoring module has scored the weights of each edge but the regularizer defined in LGRPool enforces the latent variable to be aligned with predicted latent variable which was obtained by propagation. This ensures the graph remains connected and leverages the inherent connectivity of the graph. It also ensures that the merging in the second step is consistent with global propagation. 
The propagation step could be considered as a special case of GGD and all random walks with different lengths are implicitly considered in the propagation module of expectation step. The loop of expectation-maximisation continues until it passes the convergence threshold. The final latent variable has two properties. Firstly, it ensures that the propagation module in expectation step discovers global information and the scoring and merging modules attends to all local structures at different scales of the graph. Each scale of the graph is associated to a different pooling layer but the present work only gets feedback from the pooling information of the last layer since the intermediate layers will be adjusted automatically by backpropagation. 
\subsection{Aligning Local to Global Features}
Although there are many ways to model local structures such as defining kernels in \cite{LucaCosmo1014} that learns the hidden motifs inside the graph with the same analogy to convolutional neural networks(CNN), the alignment between local information and global information is often overlooked since the higher order structure information such as relative distance of local structures is lost in these modelings. As discussed before, there are many perspective on how to define and apply global features as well. Some concatenate global positional embedding to traditional message passing methods but this alignment at different scales of graph is totally missing since there is no hierarchical modeling in most methods. \cite{MosheEliasof2024} concatenates the first and the last layer of GNN and then pass it to k multi-layer perceptron(MLP) networks to represent a global vector for label k. The loss function considers the relationship between the label and node features by providing an inductive bias that similar nodes belong to a respective label while requiring the dissimilarity of node features that do not belong to that label and its features. With the same spirit and by using top-k eigenvectors of the common graph operators, \cite{NingyuanHuang2022} concatenates k MLP networks to model all granularities from very local to very global representation but ignores the topology of the graph completely. This strong emphasis on node labels provides a domain shift between training graphs and test graphs since the topological information is missed in the learning mode and overfitting is unavoidable. To this end, we propose a HGP approach in an expectation maximisation framework, such that different layers correspond to different scales and aligns local information in each pooling layer in the maximisation step to the global information provided by the approximation of personalized page rank in the expectation step.
\subsection{Expectation step}
The expectation step consists of prediction module and the propagation module. Since the latent variable in expectation step could be seen as a prediction step, only a priori estimation would suffice it. 
\begin{equation}
\begin{split}
Z^{0} &= H \\
H &= f_{\theta}(X) \\
\end{split}
\end{equation}
Z is considered as the latent variable in the Expectation step. $f_{\theta}$ is just a fully connected neural network in the prediction module that is modeled by a neural network and is applied to all nodes of the graph. In the propagation module, the following approximation of personalized page rank is used via approximating the matrix inversion:
\begin{equation}
\begin{split}
Z^{k+1} &= (1-\alpha)\hat{\Tilde{A}}Z^{(k)}+\alpha H \\
Z^{k_{final}} &= \textbf{softmax} ((1-\alpha))\hat{\Tilde{A}}Z^{(k_{final}-1)}+\alpha H)
\end{split}
\end{equation}
where $\alpha$ is a hyperparameter.
The objective function in the expectation step for the present paper is a graph classification problem which is estimated by the following global mean pooling which is just an average of the node features:
\begin{equation}\label{eq-expectationstepLoss}
\begin{split}
y_{pred} &= \frac{1}{N} \sum_{i=1}^{N} Z^{k_{final}}_{i} \\
\mathcal{L}_{exp}&=\textbf{CrossEntropy}(y_{true}, y_{pred})
\end{split}
\end{equation}
where N is the number of nodes and $y_{true}$ are the true graph labels.
\subsection{Maximization step}
The edge scores can be obtained by the following symmetrized function to be invariant on any permutation of node's order.
\begin{equation}\label{eq:edgeScoring}
\begin{split}
s_{ij}&=\frac{1}{2}(\sigma(a[W_{pool}Z_{i}||W_{pool}Z_{j}]) \\
&+\sigma(a[W_{pool}Z_{j}||W_{pool}Z_{i}]))
\end{split}
\end{equation}
where $\sigma$ in \eqref{eq:edgeScoring} is a sigmoid function , $W_{pool}$ and $a$ are trainable matrices.
The following normalization is necessary to compute the nodes features in the coarsened graph:
\begin{equation}\label{eq:norm}
S_{norm_{ij}} = \frac{s_{ij}1_{s_{ij}\ge s_{thre}}}{\sum_{j\in N(i)} 1_{s_{ij} \ge s_{thre}}}
\end{equation}
Please note that the coarsening is done by only merging of similar nodes. 
The prediction-correction loss is defined as follows:
\begin{equation}\label{eq-pre-corLoss}
\begin{split}
\mathcal{L}_{pre-cor}&= \sum_{i\in V} ||Z_{g(i)}^{cor}-Z_{i}^{pre}||_{2}^{2} \\
&-\sum_{(g(i),g(j))\in E} ||Z_{g(i)}^{cor}-Z_{g(j)}^{cor}||_{2}^{2}
\end{split}
\end{equation}
where $Z_{g(i)}$ is the representation of the global mapping of node $i$ to node $g(i)$ and the correction and prediction Z are defined as follows:
\begin{equation}
\begin{split}
Z_{i}^{cor} &:= Z^{(l+1)}_{i}   \\
Z_{i}^{pre} &:= Z^{k_{final}}
\end{split}
\end{equation}
The first term on the right hand side of \eqref{eq-pre-corLoss} models the local global regularisation between the first layer and the layer $K-1$ is defined as the $L_2$ norm between their corresponding features and is shown in Figure~\ref{fig-anealingProcess.png}. The second term enforces the representations to stay in a compact and close regions of the state space to promote a dense representation and to avoid curse of dimensionality as much as possible.
Thus, the total loss is defined as:
\begin{equation}\label{eq:total}
\mathcal{L}_{tot} = \mathcal{L}_{exp} +\gamma \mathcal{L}_{pre-cor}
\end{equation}
where $\gamma$ is a hyperparameter and $\mathcal{L}_{exp}$ is the graph classification loss defined in \eqref{eq-expectationstepLoss} since the present work is limited to graph level tasks and not node level classification. Note that only the last layer of the final pooling layer is used to compute $\mathcal{L}_{cl}$ to avoid overfitting. Even the regularizer only uses the last layer information since other layers features are automatically updated by backpropagation. An ablation study is carried out and is shown in Table~\ref{tab:gamma} to see the effect of $\gamma$ which is a trade off between the two losses.
\section{Experiments}
\subsection{Dataset Statistics}
Since the core idea of the present paper is to focus on aligning global topological information for the task of graph classification and not on the node classification, the experiments are only done on such benchmarks.
The statistics of dataset is shown in Table~\ref{tab:dataset_statistics}. The settings of dataset such as test sets are exactly the same as \cite{FangdaGu2020}. Table~\ref{tab:hyperparameters} shows the optimized hyperparameters. $\alpha$ is the teleportation probability, k is the number of iterations in each iteration of propagation module in prediction step. Learning rate is scheduled dynamically with decaying by a factor of $0.95$ every 10 epochs. 
\subsection{Ablation Study}
An ablation study is done to see the effect of $\gamma$ on the graph classification that is shown in table~\ref{tab:gamma}. As the table shows a global minimum is formed at 0.2 for most of datasets except the protein dataset which is shifted to 0.25. It can be inferred from table~\ref{tab:gamma} that increasing $\gamma$ beyond 0.2 for most experimented datasets deteriorates the expectation loss that is responsible for the effect of global topological features on the graph classification problem. On the other hand, reducing the $\gamma$ lower than 0.2 would diminish the multiscale information which is obtained by hierarchical graph pooling. Figure~\ref{tab:graph_classification_results} shows the performance of different HGP methods for some benchmark graph classification, and as could be seen, there is just a slight improvement in only two of these four datasets.
\section{Conclusion}
To circumvent the local nature of GNN, we proposed a flexible framework which could be seen as an expectation maximization framework for HGP that aligns global topological features with local features at each scale which slightly outperforms on some graph classification benchmarks. In the expectation step, feature learning is separated from propagation which is a known technique to avoid the need for multiple layers of GNN to connect two distant nodes. In the Maximisation step, a regularizer is designed to align hierarchical graph pooling representation to the representation that is obtained in the expectation step. Since the present work proposes a framework, any known GNN model or HGP could be used in the expectation or maximisation module which can further outperform the SOTA for graph classification. In future works, we will leverage general graph random features(g-GRF) \cite{IsaacReid2024} to implicitly model structural patterns like motifs and graphlets.
\bibliographystyle{named}
\bibliography{ijcai24}

\begin{thebibliography}{}

\bibitem[\protect\citeauthoryear{Baek \bgroup \em et al.\egroup }{2021}]{JinheonBaek2021}
Jinheon Baek, Minki Kang, and Sung~Ju Hwang.
\newblock Accurate learning of graph representations with graph multiset pooling.
\newblock {\em ArXiv}, abs/2102.11533, 2021.

\bibitem[\protect\citeauthoryear{Barbero \bgroup \em et al.\egroup }{2024}]{Barbero2024}
Federico Barbero, Ameya Velingker, Amin Saberi, Michael Bronstein, and Francesco~Di Giovanni.
\newblock Locality-aware graph-rewiring in gnns.
\newblock 2024.

\bibitem[\protect\citeauthoryear{Bhowmick \bgroup \em et al.\egroup }{2024}]{Bhowmick2024}
Aritra Bhowmick, Mert Kosan, Zexi Huang, Ambuj Singh, and Sourav Medya.
\newblock Dgcluster: A neural framework for attributed graph clustering via modularity maximization.
\newblock {\em Proceedings of the AAAI Conference on Artificial Intelligence}, 38:11069--11077, 03 2024.

\bibitem[\protect\citeauthoryear{Bianchi \bgroup \em et al.\egroup }{2019}]{Bianchi2019}
Filippo~Maria Bianchi, Daniele Grattarola, and Cesare Alippi.
\newblock Spectral clustering with graph neural networks for graph pooling.
\newblock In {\em International Conference on Machine Learning}, 2019.

\bibitem[\protect\citeauthoryear{Bianchi \bgroup \em et al.\egroup }{2020}]{Bianchi2020}
Filippo~Maria Bianchi, Daniele Grattarola, and Cesare Alippi.
\newblock Spectral clustering with graph neural networks for graph pooling.
\newblock 11 2020.

\bibitem[\protect\citeauthoryear{Bojchevski \bgroup \em et al.\egroup }{2020}]{Bojchevski2020}
Aleksandar Bojchevski, Johannes Gasteiger, Bryan Perozzi, Amol Kapoor, Martin Blais, Benedek Rózemberczki, Michal Lukasik, and Stephan Günnemann.
\newblock Scaling graph neural networks with approximate pagerank.
\newblock pages 2464--2473, 08 2020.

\bibitem[\protect\citeauthoryear{Brüel-Gabrielsson \bgroup \em et al.\egroup }{2022}]{Gabrielsson2022}
Rickard Brüel-Gabrielsson, Mikhail Yurochkin, and Justin Solomon.
\newblock Rewiring with positional encodings for graph neural networks.
\newblock 01 2022.

\bibitem[\protect\citeauthoryear{Chen \bgroup \em et al.\egroup }{2020}]{DexiongChen2020}
Dexiong Chen, Laurent Jacob, and Julien Mairal.
\newblock Convolutional kernel networks for graph-structured data.
\newblock {\em ArXiv}, abs/2003.05189, 2020.

\bibitem[\protect\citeauthoryear{Chien \bgroup \em et al.\egroup }{2020}]{EliChien2020}
Eli Chien, Jianhao Peng, Pan Li, and Olgica Milenkovic.
\newblock Adaptive universal generalized pagerank graph neural network.
\newblock {\em arXiv: Learning}, 2020.

\bibitem[\protect\citeauthoryear{Choromanski \bgroup \em et al.\egroup }{2024}]{KrzysztofChoromanski2024}
Krzysztof Choromanski, Isaac Reid, Arijit Sehanobish, and Avinava Dubey.
\newblock Optimal time complexity algorithms for computing general random walk graph kernels on sparse graphs.
\newblock 10 2024.

\bibitem[\protect\citeauthoryear{Cosmo \bgroup \em et al.\egroup }{2024}]{LucaCosmo1014}
Luca Cosmo, Giorgia Minello, Alessandro Bicciato, Michael~M. Bronstein, Emanuele Rodolà, Luca Rossi, and Andrea Torsello.
\newblock Graph kernel neural networks.
\newblock {\em IEEE Transactions on Neural Networks and Learning Systems}, pages 1--14, 2024.

\bibitem[\protect\citeauthoryear{Diehl}{2019}]{Diehl2019}
Frederik Diehl.
\newblock Edge contraction pooling for graph neural networks.
\newblock 05 2019.

\bibitem[\protect\citeauthoryear{Eijkelboom \bgroup \em et al.\egroup }{2023}]{Eijkelboom2023}
Floor Eijkelboom, Erik Bekkers, Michael Bronstein, and Francesco~Di Giovanni.
\newblock Can strong structural encoding reduce the importance of message passing?
\newblock 2023.

\bibitem[\protect\citeauthoryear{Eliasof and Treister}{2024}]{MosheEliasof2024}
Moshe Eliasof and Eran Treister.
\newblock Global-local graph neural networks for node-classification.
\newblock 06 2024.

\bibitem[\protect\citeauthoryear{Eliasof \bgroup \em et al.\egroup }{2022}]{MosheEliasof2022}
Moshe Eliasof, Eldad Haber, and Eran Treister.
\newblock pathgcn: Learning general graph spatial operators from paths.
\newblock 07 2022.

\bibitem[\protect\citeauthoryear{Eliasof \bgroup \em et al.\egroup }{2023}]{MosheEliasof2023}
Moshe Eliasof, Fabrizio Frasca, Beatrice Bevilacqua, Eran Treister, Ga~Chechik, and Haggai Maron.
\newblock Graph positional encoding via random feature propagation.
\newblock In {\em Proceedings of the 40th International Conference on Machine Learning}, ICML'23. JMLR.org, 2023.

\bibitem[\protect\citeauthoryear{Feng \bgroup \em et al.\egroup }{2022}]{AosongFeng2022}
Aosong Feng, Chenyu You, Shiqiang Wang, and Leandros Tassiulas.
\newblock Kergnns: Interpretable graph neural networks with graph kernels.
\newblock {\em Proceedings of the AAAI Conference on Artificial Intelligence}, 36:6614--6622, 06 2022.

\bibitem[\protect\citeauthoryear{Galland and marc lelarge}{2021}]{Galland2021}
Alexis Galland and marc lelarge.
\newblock Graph pooling by edge cut.
\newblock 2021.

\bibitem[\protect\citeauthoryear{Gao and Ji}{2019a}]{HongyangGao2019-graphUnet}
Hongyang Gao and Shuiwang Ji.
\newblock Graph u-nets.
\newblock {\em IEEE Transactions on Pattern Analysis and Machine Intelligence}, 44:4948--4960, 2019.

\bibitem[\protect\citeauthoryear{Gao and Ji}{2019b}]{HongyangGao2019}
Hongyang Gao and Shuiwang Ji.
\newblock Graph u-nets.
\newblock In Kamalika Chaudhuri and Ruslan Salakhutdinov, editors, {\em Proceedings of the 36th International Conference on Machine Learning}, volume~97 of {\em Proceedings of Machine Learning Research}, pages 2083--2092. PMLR, 09--15 Jun 2019.

\bibitem[\protect\citeauthoryear{Gao \bgroup \em et al.\egroup }{2019}]{HongyangGao2019-gpool}
Hongyang Gao, Yongjun Chen, and Shuiwang Ji.
\newblock Learning graph pooling and hybrid convolutional operations for text representations.
\newblock pages 2743--2749, 05 2019.

\bibitem[\protect\citeauthoryear{Gasteiger \bgroup \em et al.\egroup }{2019a}]{Gasteiger2019}
Johannes Gasteiger, Aleksandar Bojchevski, and Stephan Günnemann.
\newblock Predict then propagate: Graph neural networks meet personalized pagerank.
\newblock 02 2019.

\bibitem[\protect\citeauthoryear{Gasteiger \bgroup \em et al.\egroup }{2019b}]{JohannesGasteiger2019}
Johannes Gasteiger, Stefan Weiss~enberger, and Stephan Gunnemann.
\newblock Diffusion improves graph learning.
\newblock In {\em Advances in Neural Information Processing Systems}, volume~32. Curran Associates, Inc., 2019.

\bibitem[\protect\citeauthoryear{Gu \bgroup \em et al.\egroup }{2020}]{FangdaGu2020}
Fangda Gu, Heng Chang, Wenwu Zhu, Somayeh Sojoudi, and Laurent~El Ghaoui.
\newblock Implicit graph neural networks.
\newblock {\em ArXiv}, abs/2009.06211, 2020.

\bibitem[\protect\citeauthoryear{Haddadian \bgroup \em et al.\egroup }{2024}]{Haddadian2024}
Parsa Haddadian, Roya Booryaee, Rooholah Abedian, and Ali Moeini.
\newblock Multi-hop attention-based graph pooling: A personalized pagerank perspective.
\newblock 03 2024.

\bibitem[\protect\citeauthoryear{Huang \bgroup \em et al.\egroup }{2022}]{NingyuanHuang2022}
Ningyuan Huang, Soledad Villar, Carey~E. Priebe, Da~Zheng, Cheng-Fu Huang, Lin~F. Yang, and Vladimir Braverman.
\newblock From local to global: Spectral-inspired graph neural networks.
\newblock {\em ArXiv}, abs/2209.12054, 2022.

\bibitem[\protect\citeauthoryear{Kalofolias \bgroup \em et al.\egroup }{2021}]{Kalofolias2021}
Janis Kalofolias, Pascal Welke, and Jilles Vreeken.
\newblock {\em SUSAN: The Structural Similarity Random Walk Kernel}, pages 298--306.
\newblock 04 2021.

\bibitem[\protect\citeauthoryear{Lee \bgroup \em et al.\egroup }{2018}]{JuhoLee2018}
Juho Lee, Yoonho Lee, Jungtaek Kim, Adam~R. Kosiorek, Seungjin Choi, and Yee~Whye Teh.
\newblock Set transformer.
\newblock {\em ArXiv}, abs/1810.00825, 2018.

\bibitem[\protect\citeauthoryear{Lee \bgroup \em et al.\egroup }{2019}]{JunhyunLee2019}
Junhyun Lee, Inyeop Lee, and Jaewoo Kang.
\newblock Self-attention graph pooling.
\newblock 04 2019.

\bibitem[\protect\citeauthoryear{Liu \bgroup \em et al.\egroup }{2020}]{MengLiu2020}
Meng Liu, Hongyang Gao, and Shuiwang Ji.
\newblock Towards deeper graph neural networks.
\newblock In {\em Proceedings of the 26th ACM SIGKDD International Conference on Knowledge Discovery \& Data Mining}, KDD '20, page 338–348, New York, NY, USA, 2020. Association for Computing Machinery.

\bibitem[\protect\citeauthoryear{Liu \bgroup \em et al.\egroup }{2022}]{YueLiu2022}
Yue Liu, Lixin Cui, Yue Wang, and Lu~Bai.
\newblock Abdpool: Attention-based differentiable pooling.
\newblock In {\em 2022 26th International Conference on Pattern Recognition (ICPR)}, pages 3021--3026, 2022.

\bibitem[\protect\citeauthoryear{Long \bgroup \em et al.\egroup }{2021}]{QingqingLong2021}
Qingqing Long, Yilun Jin, Yi~Wu, and Guojie Song.
\newblock Theoretically improving graph neural networks via anonymous walk graph kernels.
\newblock pages 1204--1214, 04 2021.

\bibitem[\protect\citeauthoryear{Ma \bgroup \em et al.\egroup }{2024}]{LihengMa2024}
Liheng Ma, Soumyasundar Pal, Yitian Zhang, Jiaming Zhou, Yingxue Zhang, and Mark Coates.
\newblock Ckgconv: General graph convolution with continuous kernels.
\newblock 07 2024.

\bibitem[\protect\citeauthoryear{Mingxing \bgroup \em et al.\egroup }{2022}]{XuMingxing2022}
Xu~Mingxing, Wenrui Dai, Chenglin Li, Junni Zou, and Hongkai Xiong.
\newblock Liftpool: Lifting-based graph pooling for hierarchical graph representation learning.
\newblock 04 2022.

\bibitem[\protect\citeauthoryear{Nikolentzos and Vazirgiannis}{2020}]{Nikolentzos2020}
Giannis Nikolentzos and Michalis Vazirgiannis.
\newblock Random walk graph neural networks.
\newblock In H.~Larochelle, M.~Ranzato, R.~Hadsell, M.F. Balcan, and H.~Lin, editors, {\em Advances in Neural Information Processing Systems}, volume~33, pages 16211--16222. Curran Associates, Inc., 2020.

\bibitem[\protect\citeauthoryear{Reid \bgroup \em et al.\egroup }{2023}]{IsaacReid2023}
Isaac Reid, Krzysztof Choromanski, Eli Berger, and Adrian Weller.
\newblock General graph random features.
\newblock In {\em International Conference on Learning Representations}, 2023.

\bibitem[\protect\citeauthoryear{Reid \bgroup \em et al.\egroup }{2024}]{IsaacReid2024}
Isaac Reid, Krzysztof~Marcin Choromanski, Eli Berger, and Adrian Weller.
\newblock General graph random features.
\newblock In {\em The Twelfth International Conference on Learning Representations}, 2024.

\bibitem[\protect\citeauthoryear{Roth and Liebig}{2022}]{Roth2022}
Andreas Roth and Thomas Liebig.
\newblock Transforming pagerank into an infinite-depth graph neural network.
\newblock In {\em ECML/PKDD}, 2022.

\bibitem[\protect\citeauthoryear{Shen \bgroup \em et al.\egroup }{2024}]{XuShen2024}
Xu~Shen, Pietro Liò, Lintao Yang, Ru~Yuan, Yuyang Zhang, and Chengbin Peng.
\newblock Graph rewiring and preprocessing for graph neural networks based on effective resistance.
\newblock {\em IEEE Transactions on Knowledge and Data Engineering}, 36(11):6330--6343, 2024.

\bibitem[\protect\citeauthoryear{Snelleman \bgroup \em et al.\egroup }{2024}]{Snelleman2024}
T.~Snelleman, B.~Renting, H.~Hoos, and J.~Rijn.
\newblock Edge-based graph component pooling.
\newblock 09 2024.

\bibitem[\protect\citeauthoryear{Stanovic \bgroup \em et al.\egroup }{2025}]{Stanovic2025}
Stevan Stanovic, Benoit Gaüzère, and Luc Brun.
\newblock Graph neural networks with maximal independent set-based pooling: Mitigating over-smoothing and over-squashing.
\newblock {\em Pattern Recognition Letters}, 187:14--20, 2025.

\bibitem[\protect\citeauthoryear{Tsitsulin \bgroup \em et al.\egroup }{2024}]{Tsitsulin2024}
Anton Tsitsulin, John Palowitch, Bryan Perozzi, and Emmanuel M\"{u}ller.
\newblock Graph clustering with graph neural networks.
\newblock {\em J. Mach. Learn. Res.}, 24(1), March 2024.

\bibitem[\protect\citeauthoryear{Wimalawarne and Suzuki}{2021}]{Wimalawarne2021}
Kishan Wimalawarne and Taiji Suzuki.
\newblock Layer-wise adaptive graph convolution networks using generalized pagerank.
\newblock In {\em Asian Conference on Machine Learning}, 2021.

\bibitem[\protect\citeauthoryear{Wu \bgroup \em et al.\egroup }{2022}]{WuJunran2022}
Junran Wu, Xueyuan Chen, Ke~Xu, and Shangzhe Li.
\newblock Structural entropy guided graph hierarchical pooling.
\newblock In {\em International Conference on Machine Learning}, 2022.

\bibitem[\protect\citeauthoryear{Xu \bgroup \em et al.\egroup }{2024}]{XuZhuo2024}
Zhuo Xu, Lixin Cui, Ming Li, Yue Wang, Ziyu Lyu, Hangyuan Du, Lu~Bai, Philip~S. Yu, and Edwin~R. Hancock.
\newblock Sshpool: The separated subgraph-based hierarchical pooling.
\newblock 2024.

\bibitem[\protect\citeauthoryear{Ying \bgroup \em et al.\egroup }{2018}]{RexYing2018}
Rex Ying, Jiaxuan You, Christopher Morris, Xiang Ren, William~L. Hamilton, and Jure Leskovec.
\newblock Hierarchical graph representation learning with differentiable pooling.
\newblock {\em ArXiv}, 2018.

\bibitem[\protect\citeauthoryear{Yu \bgroup \em et al.\egroup }{2024}]{YuZhizhi2024}
Zhizhi Yu, Bin Feng, Dongxiao He, Zizhen Wang, Yuxiao Huang, and Zhiyong Feng.
\newblock Lg-gnn: Local-global adaptive graph neural network for modeling both homophily and heterophily.
\newblock In Kate Larson, editor, {\em Proceedings of the Thirty-Third International Joint Conference on Artificial Intelligence, {IJCAI-24}}, pages 2515--2523. International Joint Conferences on Artificial Intelligence Organization, 8 2024.
\newblock Main Track.

\bibitem[\protect\citeauthoryear{Zhang \bgroup \em et al.\egroup }{2018}]{ZhenZhang2018}
Zhen Zhang, Mianzhi Wang, Yijian Xiang, and Yan Huang.
\newblock Retgk: Graph kernels based on return probabilities of random walks.
\newblock 09 2018.

\bibitem[\protect\citeauthoryear{Zhao \bgroup \em et al.\egroup }{2024}]{ZhehanZhao2024}
Zhehan Zhao, Lu~Bai, Lixin Cui, Ming Li, Yue Wang, Lixiang Xu, and Edwin Hancock.
\newblock Enadpool: The edge-node attention-based differentiable pooling for graph neural networks.
\newblock 05 2024.

\end{thebibliography}
\end{document}